\documentclass[letterpaper, 10 pt, conference]{ieeeconf}  

\IEEEoverridecommandlockouts                              

\usepackage{todonotes} 
\usepackage{cite}
\expandafter\let\csname proof\endcsname\relax
\expandafter\let\csname endproof\endcsname\relax
\usepackage{amsmath,amssymb,amsfonts}
\usepackage{graphicx}
\usepackage{textcomp}
\usepackage{xcolor}
\usepackage{amsthm}
\usepackage{hyperref}
\usepackage[ruled]{algorithm2e} 
\usepackage{algorithmic}
\usepackage{booktabs}
\usepackage{balance}
\usepackage{gensymb}
\def\BibTeX{{\rm B\kern-.05em{\sc i\kern-.025em b}\kern-.08em
		T\kern-.1667em\lower.7ex\hbox{E}\kern-.125emX}}
\newtheorem{theorem}{Theorem}
\newtheorem{corollary}{Corollary}[theorem]
\newtheorem{lemma}{Lemma}

\newtheorem*{remark}{Remark}
\newtheorem{definition}{Definition}

\usepackage{threeparttable}
\title{\LARGE \bf
	Safety-Critical Optimal Control for Robotic Manipulators 
	
	in A Cluttered Environment
}

\author{Xuda Ding$^{1}$, Han Wang$^{2}$, Yi Ren$^{3}$, Yu Zheng$^{3}$, Cailian Chen$^{1}$, Jianping He$^{1}$
	\thanks{$^1$: the Dept. of Automation, Shanghai Jiao Tong University, Shanghai, China. email: {\tt\small\{dingxuda, cailianchen, jphe\}@sjtu.edu.cn}.
	}
	\thanks{$^2$: the Dept. of Engineering Science, University of Oxford, Oxford, United Kingdom. E-mails: {\tt\small han.wang@eng.ox.ac.uk}}
	\thanks{$^3$: the Tencent Robotics X Lab, Shenzhen, China. email: {\tt\small \{evanyren, petezheng\}@tencent.com, yren.tum@outlook.com}}
}

\begin{document}

	\maketitle
	\thispagestyle{empty}
	\pagestyle{empty}

	\begin{abstract}
		Designing safety-critical control for robotic manipulators is challenging, especially in a cluttered environment.
		First, the actual trajectory of a manipulator might deviate from the planned one due to the complex collision environments and non-trivial dynamics, leading to collision;
		Second, the feasible space for the manipulator is hard to obtain since the explicit distance functions between collision meshes are unknown.
		By analyzing the relationship between the safe set and the controlled invariant set, this paper proposes a data-driven control barrier function (CBF) construction method, which extracts CBF from distance samples.
		Specifically, the CBF guarantees the controlled invariant property for considering the system dynamics.
		The data-driven method samples the distance function and determines the safe set.
		Then, the CBF is synthesized based on the safe set by a scenario-based sum-of-square
		(SOS) program.
		Unlike most existing linearization-based approaches, our method reserves the volume of the feasible space for planning without approximation, which helps find a solution in a cluttered environment.
		The control law is obtained by solving a CBF-based quadratic program in real time, which works as a safe filter for the desired planning-based controller.
		Moreover, our method guarantees safety with the proven probabilistic result.
		Our method is validated on a 7-DOF manipulator in both real and virtual cluttered environments. 
		The experiments show that the manipulator is able to execute tasks where the clearance between obstacles is in millimeters.
	\end{abstract}

	\section{Introduction}
	Safety-critical motion planning is fundamental for applications of robotic manipulators since manipulators need to be driven to a specified goal without collisions \cite{thakar2022manipulator,rybus2020point}.
	The whole body must have no collisions with obstacles and itself.
	In the past decades, path planning methods combined with tracking control have been proposed to generate a safe path for manipulators.
	
	Encouraged by the efficient applications of search methods in path planning, rapidly-exploring random trees \cite{rybus2020point} and probabilistic road maps  \cite{kavraki1996probabilistic} are proposed.
	These methods are probabilistically complete, which means they may take a long time to achieve asymptotic optimality.
	To deal with the optimality problem, the non-linear optimization algorithms such as TrajOpt \cite{schulman2014motion}, CHOMP \cite{ratliff2009chomp}, and OMPL \cite{sucan2012open} are used to plan joint trajectories.
	However, the actual trajectory might deviate from the planned one due to the complex dynamics, and cluttered environments \cite{singletary2022safety}.
	A slight trajectory deviation might cause a collision in a cluttered environment where the potential clearance between the manipulator and obstacles is less than a few millimeters, lead to unsafe.
	
	\begin{figure}[t]
		\centering
		\includegraphics[width=1\linewidth]{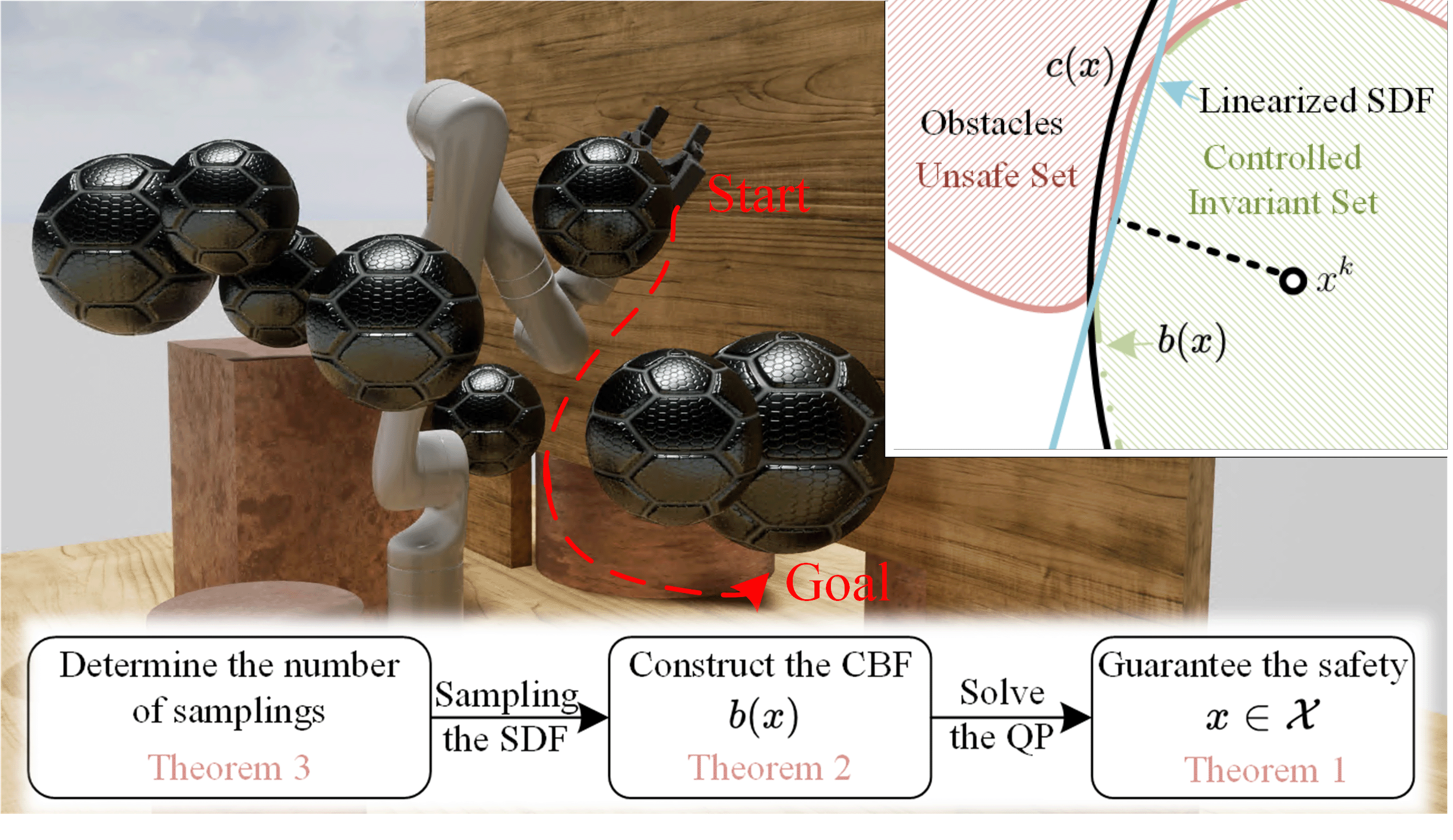}
		\caption{The illustration of the proposed safety-critical optimal control method. The linearized SDF at $x^k$ overlooks some feasible space. Our approach uses SDF samples to construct the control invariant set and reserves the volume of the feasible space.
			The subfigure on the right side is the illustration of the CBF construction.
			The subfigure at the bottom shows the control procedure.
		}
		\label{fig:front}
				\vspace{-20pt}
	\end{figure}
	
	System-dynamic-based optimization methods are proposed to deal with deviation problems.
	The system dynamic is considered in the constraints of the optimization.
	The constraints also amend that the manipulator has no collision with obstacles.
	In this way, motion planning can satisfy the collision-free requirement.
	Mixed-integer linear problems and mixed-integer quadratic problems are proposed to generate the motion plan by sampling and calculating the points on the manipulator's surface to the obstacles \cite{ding2011mixed,yin2004hybrid}.
	Since the number of integer variables constraints is large, these methods take seconds to minutes to solve.
	The Quadratic Problem (QP) approach is used for motion plans in an online manner without integer variables \cite{zhang2020optimization,pankert2020perceptive}.
	
	In recent, with the revisiting of the controlled invariant set and Control Barrier Function (CBF) in control theory, the CBF-based QP approach is used for safe-critical motion planning and enforces the safety \cite{ames2016control,singletary2022safety}.
	However, it is still challenging to construct constraint functions concerning collision since the manipulator's configuration space does not match the obstacles' space.
	Sampling-based methods are mainstream, which are proposed to obtain the Signed Distance Function (SDF) between two objects considering the meshes \cite{schulman2014motion,gilbert1988fast,van2001proximity}.
	The SDF represents the shortest distance and gives the nearest points of two objects in the manipulators' configuration space.
	The differentiation of SDF is discontinuous since there is a min and max operation to obtain the function \cite{schulman2014motion}.
	Directly using SDF in QP can lead to a local minimum and disobey the CBF-based optimization requirement of continuity \cite{singletary2022safety}.
	Linearization of the SDF is used to solve this problem, and then the constraint is constructed \cite{schulman2014motion,zhang2020optimization,singletary2022safety}.
	However, the linearization of SDF leads to a small feasible space for optimization and an infeasible solution, especially when the collision meshes are complex envelopes in a cluttered environment \cite{schulman2014motion}.
	Furthermore, the linearization of SDF or SDF gives a determination for safety, but they do not necessarily equal CBF since neither guarantees control invariance.
	
	Overall, it is still challenging to design a safety-critical control method for manipulators working in a cluttered environment.
	This letter aims to design an optimal control method in a cluttered environment and enforces safety based on the CBF.
	Some critical issues need to be solved
	i) Since the explicit SDF cannot be determined, the safe set is unknown.
	ii) How to determine the safe set based on SDF sampling and extract a candidate CBF based on a safe set.
	iii) The number of SDF samples needed for estimating the safe set and extracting the CBF, the probability of safety still needs to be determined.
	Inspired by the scenario approach and CBF construction \cite{campi2008exact,wang2022safe}, we found it is possible to solve these challenges.
	Our approach first formulates the safe set based on SDF, considering the collisions with obstacles and the manipulator itself.
	Then, we analyze the relationship between the safe set and the controlled invariant set.
	The relationship provides the basic principle for CBF construction.
	The ellipsoidal-Lyapunov-like CBF candidate is extracted based on the safe set with online sampling, which guarantees that the control invariant set is a subset of the safe set.
	Furthermore, a theoretical probability guarantee of safety is given based on the number of SDF sampling used to extract the CBF.
	Fig.\ref{fig:front} illustrates the proposed safety-critical optimal control method for manipulators in a cluttered environment.
	
	The main contributions of this letter are
	\begin{itemize}
		\item A novel data-driven CBF construction methodology is proposed. 
		The novelty lies in: i) The relationship between SDF and CBF is explored.
		By solving a scenario-based sum-of-square (SOS) program, CBF is synthesized based on the dynamic model and SDF samples.
		ii) A maximum optimization problem is used to find the CBF parameters without linearizing the SDF, which reserves the feasible space for planning.
		\item Unsafety probability is analytically expressed based on the number of samples and support constraints.
		The probability guides the users to choose the SDF sampling number for CBF construction and control.
		\item The proposed approach is implemented on a full-scale manipulator (Kinova Gen3) in an obstacle-cluttered environment.
		The speed and efficacy of this method are extensively explored in real-world environments, and the method has been shown to be real-time and safe.
	\end{itemize}
	
	The rest of this paper is organized as follows.
	Section \ref{sec:pre} formulates the optimal control problem with CBF and introduces the relationship among safety, control invariant set, and the SDF.
	The data-driven CBF extraction method and the probability guarantee for safety are given in Section \ref{sec:CBF}.
	Simulation and implementation are shown in Section \ref{sec:sim}.
	Section \ref{sec:con} concludes the paper.
	
	\section{Preliminaries}\label{sec:pre}
	\subsection{System Dynamics and Objective}
	Consider a dynamic model of robotic manipulator
	\begin{equation}\label{eq:kineticmodel}
		\dot{\boldsymbol{x}} = f(\boldsymbol{x}, \boldsymbol{u}),
	\end{equation}
	where state $\boldsymbol{x}(t) \in \boldsymbol{\mathcal{X}} \subset \mathbb{R}^n$ and input $\boldsymbol{u}(t) \in \boldsymbol{\mathcal{U}} \subset \mathbb{R}^m$, $\boldsymbol{\mathcal{U}}$ is a compact set and $f(\cdot,\cdot)$ is Lipschitz continuous.
	Our goal is to find a control policy that allows the system to navigate the manipulator $\mathcal{A}$ from the start position to the goal while optimizing an objective function and avoiding collision with obstacles $\mathcal{O}_1$, $\mathcal{O}_2$, $\dots$, $\mathcal{O}_M \subset \mathbb{R}^3$, where $M \in \mathbb{Z}^+$.
	
	To explore intrinsic safety, we first need to find a set $\boldsymbol{\mathcal{B}}$, where the state of the manipulator $\mathcal{A}$ is always kept within.
	Such set $\boldsymbol{\mathcal{B}}$ is a \textit{Controlled Invariant Set}.
	\begin{definition}[Controlled Invariant Set]
		The set $\boldsymbol{\mathcal{B}}$ is a controlled invariant set for system \eqref{eq:kineticmodel} if for every $\boldsymbol{x}_0 \in \boldsymbol{\mathcal{B}}$, there exists an input $\boldsymbol{u}(t)$ such that $\boldsymbol{x}(t) \in \boldsymbol{\mathcal{B}}$ for time $t \in [0, t_{\max})$. When $t_{\max} = \infty$, $f(\cdot,\cdot)$ is \emph{forward complete}.
	\end{definition}
	The \textit{Controlled Invariant Set} guarantees the states of the controlled object flow in the region with respect to the system dynamics \cite{korda2014convex}.
	Roughly speaking, the states must be in the set $\boldsymbol{\mathcal{B}}$ and never leave.
	The CBF is denoted as $b(\boldsymbol{x}):\mathbb{R}^n\to\mathbb{R}$.
	The corresponding controlled invariant set $\boldsymbol{\mathcal{B}}$ is the zero super-level set of $b(\boldsymbol{x})$.
	The CBF $b(\boldsymbol{x})$ can be incorporated into a QP to synthesize point-wise optimal control law \cite{ames2016control,singletary2022safety} and guarantee the state $\boldsymbol{x}$ is in the controlled invariant set $\boldsymbol{\mathcal{B}}$.
	\begin{equation}\tag{SCB-QP}\label{eq:SCB-QP}
		\begin{split}
			\boldsymbol{u}^*(\boldsymbol{x}) = \arg\min_u ~&\frac{1}{2}\|\boldsymbol{u}-\boldsymbol{u}_{\mathrm{des}}(\boldsymbol{x})\|_2^2\\
			\mathrm{subject~to}~&\frac{\partial b(\boldsymbol{x})}{\partial \boldsymbol{x}}f(\boldsymbol{x},\boldsymbol{u}) + \lambda b(\boldsymbol{x})\geq 0,
		\end{split}
	\end{equation}
	where $\boldsymbol{u}_\mathrm{des}(\boldsymbol{x})$ is a desired but not necessarily safe input from a predefined trajectory \cite{singletary2022safety}. $\boldsymbol{u}^*(\boldsymbol{x})$ is the filtered input, and $\lambda$ is a relaxation coefficient to reduce restriction in \eqref{eq:SCB-QP}.
	The constraint in \eqref{eq:SCB-QP} makes the filtered input $\boldsymbol{u}^*(\boldsymbol{x})$ drive the system state in the controlled invariant set $\boldsymbol{\mathcal{B}}$.
	\eqref{eq:SCB-QP} can be solved in real-time for non-linear systems.
	However, the safety of the \textit{controlled invariant set} $\boldsymbol{\mathcal{B}}$ and filtered input $\boldsymbol{u}^*$ are still unclear.
	
	\subsection{Safety and SDF}
	Here we define the \textit{safe set} to determine the safety of the \textit{controlled invariant set} $\boldsymbol{\mathcal{B}}$.
	\begin{definition}[Safe Set]
		To guarantee the manipulator is safe without collision, the states $\{\boldsymbol{x}(t)\}$ should always be in the collision-free set.
		Suppose there exists a series of functions $h_i:\mathbb{R}^n\to \mathbb{R}$:
		\begin{equation}\label{barrier}
			\begin{aligned}
				\boldsymbol{\mathcal{X}}&=\cap_{i=1}^{\mathcal{I}} {\{\boldsymbol{x}\in\mathbb{R}^n|h_i(\boldsymbol{x})\ge 0\}}, \\
				\text{Int}(\boldsymbol{\mathcal{X}})&=\cap_{i=1}^{\mathcal{I}} {\{\boldsymbol{x}\in\mathbb{R}^n|h_i(\boldsymbol{x})> 0\}},\\
				\boldsymbol{\bar{\mathcal{X}}}&=\cup_{i=1}^{\mathcal{I}} {\{\boldsymbol{x}\in\mathbb{R}^n|h_i(\boldsymbol{x})< 0\}},
			\end{aligned}
		\end{equation}
		where Int($\boldsymbol{\mathcal{X}}$) and ${\boldsymbol{\bar{\mathcal{X}}}}$ are the interior and complementary set of $\boldsymbol{\mathcal{X}}$, respectively. The sequence of functions $h_1(\boldsymbol{x}),\ldots, h_\mathcal{I}(\boldsymbol{x})$ describe the safe set, and represent the obstacles which are indexed by $1,\ldots,\mathcal{I}$.
	\end{definition}
	
	When $\boldsymbol{\mathcal{B}}\subseteq\boldsymbol{\mathcal{X}}$, the \textit{controlled invariant set} $\boldsymbol{\mathcal{B}}$ is safe, and the filtered input $\boldsymbol{u}^*$ is safe.
	On the contrary, $\boldsymbol{u}^*$ cannot guarantee safety when only a subset of $\boldsymbol{\mathcal{B}}$ is in the safe set.
	Therefore, the existence of such controller $\boldsymbol{u}^*$ is promised by the following lemma.
	\begin{lemma}\label{lem:invariance}
		There exists $\boldsymbol{u}^*$ such that system \eqref{eq:kineticmodel} is able to maintain safety under $\boldsymbol{\mathcal{X}}$, if and only if there exists a controlled invariant set $\boldsymbol{\mathcal{B}}\subseteq\boldsymbol{\mathcal{X}}$. 
	\end{lemma}
	
	\begin{remark}
		It should be noted that the function $h(\boldsymbol{x})$ used for establishing a safe set is not necessarily the same as the CBF $b(\boldsymbol{x})$, for the safe set constructed based on the obstacles may not guarantee the invariance property with respect to dynamics \eqref{eq:kineticmodel}.
	\end{remark}
	
	The meshes of the obstacles and the manipulator are considered to construct the safe set $\boldsymbol{\mathcal{X}}$ and $h_i(\boldsymbol{x})$.
	When the global information about the obstacles is known, the SDF $\text{sd}(\mathcal{A},\mathcal{O}_i)$ between manipulator $\mathcal{A}$ and the obstacle $\mathcal{O}_i$ can be obtained based on a popular way\cite{zheng2015generalized,zheng2020calculating,schulman2014motion}
	\begin{subequations}\label{eq:signed_dis}
		\begin{align}
			\text{dist}(\mathcal{A},\mathcal{O}_i) &: = \min_{\boldsymbol{d}}\{\|\boldsymbol{d}\|:(\mathcal{A}+\boldsymbol{d})\cap \mathcal{O}_i \ne \emptyset \},\label{eq:distance}\\
			\text{pen}(\mathcal{A},\mathcal{O}_i) &: = \min_{\boldsymbol{d}}\{\|\boldsymbol{d}\|:(\mathcal{A}+\boldsymbol{d})\cap \mathcal{O}_i = \emptyset \},\label{eq:penaltration}\\
			\text{sd}(\mathcal{A},\mathcal{O}_i) &: = \text{dist}(\mathcal{A},\mathcal{O}_i) - \text{pen}(\mathcal{A},\mathcal{O}_i),\label{eq:signeddist}
		\end{align}
	\end{subequations}
	where $\boldsymbol{d}$ is the posture of the manipulator.
	The safe set is determined with $h_i(\boldsymbol{x}) = \text{sd}(\mathcal{A},\mathcal{O}_i)$.
	When global information cannot be obtained, the obstacles need to be sampled and constructed.
	\textit{Voxblox} \cite{oleynikova2017voxblox} and \textit{FIESTA}\cite{han2019fiesta} can be used to incrementally construct the Euclidian Signed Distance Fields (ESDF) of the obstacles and the signed distance $\text{sd}(\mathcal{A},\mathcal{O}_i)$ is obtained.
	Then, two points $p_\mathcal{A}$, $p_{\mathcal{O}_i}$ on the manipulator and the obstacle corresponding to \eqref{eq:signed_dis} are obtained.
	Note that $p_\mathcal{A}$, $p_{\mathcal{O}_i}$ are points expressed in their own local coordinates, and they are needed to be transferred into world coordinates. 
	These two points in the world coordinates are $F_\mathcal{A}^w(\boldsymbol{x})p_\mathcal{A} \in \mathbb{R}^3$ and $F_{\mathcal{O}_i}^wp_{\mathcal{O}_i}\in \mathbb{R}^3$. The function $F^w$ is forward kinematics, which gives the pose of the manipulator and the obstacles in the world frame.
	The SDF is nonsmooth, and its explicit form is hard to obtain. 
	Linearization of SDF is used for convex formulation in \cite{schulman2014motion,pankert2020perceptive,singletary2022safety}. In this work, we propose to extract the SDF based on samplings directly.
	
	Overall, we aim to extract a safe-critical CBF based on SDF sampling to guarantee collision-free manipulation while considering the system dynamics intrinsically.
	
	\section{Dynamics-aware CBF Extraction from SDF Sampling}\label{sec:CBF}
	
	This section shows the main results of our CBF construction method based on SDF sampling.
	Our approach is data-driven since the explicit form of SDF cannot be obtained. 
	This section is organized into three parts. In Section \ref{subsec:safeset}, we formulate the \textit{safe set} based on SDF. In Section \ref{sec:cbfcon}, the CBF construction method with the SOS program is proposed. Based on the samples, a data-driven CBF construction method is given with the probabilistic guarantee for safety in Section \ref{sec:datacbf}.

	\subsection{Safe Set Construction based on SDF}\label{subsec:safeset}
	\subsubsection{Outer-collision}
	The direct interpretation of collision-free between the manipulator $\mathcal{A}$ and environmental obstacles $\{\mathcal{O}_i\}$ is $\mathrm{sd}(\mathcal{A},\mathcal{O}_i)\ge 0, \forall i \in \mathcal{I}$. However, this condition is intractable to use as a constraint in motion planning optimization problems since it has no explicit expression. 
	Formally, the outer-SDF is defined as:
	\begin{equation}\label{eq:outerdistance}
		\mathrm{sd}_{\mathrm{out}}(\boldsymbol{x}) = \max_{\|\hat{\boldsymbol{n}}_{\mathrm{out}}\|_2 = 1}\min_{p_\mathcal{A} \in \mathcal{A}\atop p_{\mathcal{O}_i} \in \mathcal{O}_i}\hat{\boldsymbol{n}}_{\mathrm{out}} \cdot \left ( F_\mathcal{A}^w(\boldsymbol{x})p_\mathcal{A} - F_{\mathcal{O}_i}^w(\boldsymbol{x})p_{\mathcal{O}_i} \right ),
	\end{equation}
	where $\hat{\boldsymbol{n}}_{\mathrm{out}}$ is the direction of the minimal translation $\boldsymbol{d}$ in \eqref{eq:signed_dis}.
	$\mathrm{sd}_{\mathrm{out}}(\boldsymbol{x})$ can be obtained by sampling points on the controlled object and obstacles, and hereafter using the GJK \cite{gilbert1988fast} or EPA \cite{van2001proximity} algorithm. With the amount of data in the magnitude of hectobit, the function can be constructed implicitly within milliseconds.

	\subsubsection{Inner-collision}
	In addition to the outer-collision scenarios considered in the last subsection, another possible collision scenario happens for different links of the manipulator. For this case, the inner-SDF is defined by:
	\begin{equation}\label{eq:innerdistance}
		\mathrm{sd}_{\mathrm{in}}(\boldsymbol{x})=\max_{\|\hat{\boldsymbol{n}}_{\mathrm{in}}\|_2 = 1}\min_{p_\mathcal{A} \in \mathcal{A}\atop p_\mathcal{A}' \in \mathcal{A}}\hat{\boldsymbol{n}}_{\mathrm{in}} \cdot \left ( F_\mathcal{A}^w(\boldsymbol{x})p_\mathcal{A} - F_\mathcal{A}^w(\boldsymbol{x})p_\mathcal{A}' \right ),
	\end{equation}
	where $p_\mathcal{A}$ and $p_\mathcal{A}'$ are different points on the different joints of the manipulator. 
	We define the overall-SDF as
	\begin{equation}
		\mathrm{sd}_{\mathrm{ov}}(\boldsymbol{x})=\min\{\mathrm{sd}_{\mathrm{out}}(\boldsymbol{x}),\mathrm{sd}_{\mathrm{in}}(\boldsymbol{x})\},
	\end{equation}
	where the composition relationship in $\mathrm{sd}_{\mathrm{ov}}(\boldsymbol{x})$ is captured by a $ \wedge $ quantifier. If for a $\boldsymbol{x}$, the manipulator is both inner-collision free and outer-collision free, then $\mathrm{sd}_{\mathrm{in}}(\boldsymbol{x})\ge 0\wedge \mathrm{sd}_{\mathrm{out}}(\boldsymbol{x})\ge 0$, which is equivalent to $\mathrm{sd}_{\mathrm{ov}}(\boldsymbol{x})\ge0$.
	When $\mathrm{sd}_{\mathrm{ov}}(\boldsymbol{x}) = 0$, the manipulator is at the boundary of collision; When $\mathrm{sd}_{\mathrm{ov}}(\boldsymbol{x}) > 0$, the manipulator is away from collision; When $\mathrm{sd}_{\mathrm{ov}}(\boldsymbol{x}) < 0$, the collision happens.
	Then, the safe set of the manipulator is constructed as the following result.
	\begin{theorem}
		Given the overall-SDF $\mathrm{sd}_{\mathrm{ov}}(\boldsymbol{x})$. The \emph{safe set} $\boldsymbol{\mathcal{X}}$ of the manipulator is:
		\begin{equation}\label{eq:safeset}
			\boldsymbol{\mathcal{X}}:=\{\boldsymbol{x}|\mathrm{sd}_{\mathrm{ov}}(\boldsymbol{x})\ge 0\}.
		\end{equation}
	\end{theorem}
	So far, the safe set is constructed based on overall-SDF.
	Then, the CBF is constructed considering the safe set and dynamic model.
	
	\subsection{CBF Construction}
	\label{sec:cbfcon}
	In most of the existing literature, the SDF is directly used or linearized and then used as a CBF in safety-critical controller design problems\cite{thirugnanam2022safety, singletary2022safety, pankert2020perceptive}. However, the safe set $\boldsymbol{\mathcal{X}}$ defined by the zero-super level set of distance function is not necessary to be a CBF candidate, as it has no controlled invariance property. 
	Actually, the whole construction procedure in Section \ref{subsec:safeset} does not utilize the manipulator's dynamic model \eqref{eq:kineticmodel}. In this subsection, we consider the dynamics and show how to synthesize a CBF $b(\boldsymbol{x})$ from the safe set $\boldsymbol{\mathcal{X}}$ based on overall-SDF $\mathrm{sd}_{\mathrm{ov}}(\boldsymbol{x})$.
	
	\begin{lemma}
		For manipulator $\mathcal{A}$ with dynamics \eqref{eq:kineticmodel}, and overall-SDF $\mathrm{sd}_{\mathrm{ov}}(\boldsymbol{x})$, $b(\boldsymbol{x})$ is a CBF candidate if
		\begin{subequations}\label{eq:cbf}
			\begin{align}
				&b(\boldsymbol{x})\le \mathrm{sd}_{\mathrm{ov}}(\boldsymbol{x}),\label{eq:positivity}\\
				\forall \boldsymbol{x}\in\partial \boldsymbol{\mathcal{B}},\exists& \boldsymbol{u}\in\boldsymbol{\mathcal{U}},~~~~\frac{\partial b(\boldsymbol{x})}{\partial \boldsymbol{x}}f(\boldsymbol{x},\boldsymbol{u})\ge 0.\label{eq:boundary}
			\end{align}
		\end{subequations}
		where $\frac{\partial b(\boldsymbol{x})}{\partial \boldsymbol{x}}f(\boldsymbol{x},\boldsymbol{u})$ is the derivative of $b(\boldsymbol{x})$ along system \eqref{eq:kineticmodel}.
	\end{lemma}
	\begin{proof}
		The right-hand side inequality in \eqref{eq:positivity} stands for that the CBF $b(\boldsymbol{x})$ should be a lower-envelope of the overall-distance function $\mathrm{sd}_{\mathrm{ov}}(\boldsymbol{x}).$ Under this property, if $b(\boldsymbol{x})\ge0$ for any $\boldsymbol{x}$ on the motion trajectory then $\mathrm{sd}_{\mathrm{ov}}(\boldsymbol{x})\ge0$, and safety can be guaranteed. 
		The second Condition \eqref{eq:boundary} is a standard controlled invariance condition. Details about this condition can be found in \cite{wang2022safety}.
	\end{proof}
	
	Without an additional objective, the construction of $b(\boldsymbol{x})$ leads to the following feasibility optimization problem.
	\begin{equation}\label{eq:feascbf}
		\mathrm{find}~b(\boldsymbol{x}),\mathrm{s.t.}~\eqref{eq:cbf}.  
	\end{equation}
	
	Although the conditions are elegant in terms of algebraic structure, it is still very hard to construct a CBF $b(\boldsymbol{x})$ by solving \eqref{eq:feascbf}. The challenges here are threefold: i) Conditions $\eqref{eq:positivity}$ and $\eqref{eq:boundary}$ should hold for infinite $\boldsymbol{x}$, which renders \eqref{eq:feascbf} to be an infinitely-constrained optimization problem; ii) $\mathrm{sd}_{\mathrm{ov}}(\boldsymbol{x})$ has only an implicit form since it is a composition of solutions to two optimization problems; iii) The constraint \eqref{eq:boundary} is especially non-convex due to the existence of control input $\boldsymbol{u}$. 
	All these three challenges will be tackled in this section, but we want to point out that the computational complexity of \eqref{eq:feascbf} is still high. The reason is that the explicit $\mathrm{sd}_{\mathrm{ov}}(\boldsymbol{x})$ cannot be obtained.
	Estimation of $\mathrm{sd}_{\mathrm{ov}}(\boldsymbol{x})$ based on a large number of samplings is time-consuming.
	The real-time computing problem will be fixed with a relaxation method in the Subsection \ref{sec:datacbf}.
	Now, we will determine the reachable set $\boldsymbol{\mathcal{C}}$ where the samples are obtained to construct CBF $b(\boldsymbol{x})$.
	\subsubsection{Reachable set determination}
	We suppose that the current state of the manipulator is $\boldsymbol{x}^k$. Given that the safe set $\boldsymbol{\mathcal{X}}$ is constructed from the sampled data online, the CBF $b(\boldsymbol{x})$ should also be synthesized online. The definition domain $\boldsymbol{\mathcal{C}}$ is therefore varying with the state $\boldsymbol{x}$. Starting from this point, the maximum movement $||\delta \boldsymbol{x}||_{\boldsymbol{x}=\boldsymbol{x}^k}$ of the manipulator is:
	\begin{equation}\label{eq:interval}
		||\delta \boldsymbol{x}||_{\boldsymbol{x}=\boldsymbol{x}^k}=\max\limits_{\boldsymbol{u}\in\boldsymbol{\mathcal{U}}} ||f(\boldsymbol{x}^k,\boldsymbol{u})dt||^2.
	\end{equation}
	The set $\boldsymbol{\mathcal{C}}$ at state $\boldsymbol{x}^k$ is then defined by a ball $\mathcal{BA}(\boldsymbol{x}^k,||\delta \boldsymbol{x}||_{\boldsymbol{x}=\boldsymbol{x}^k})$ centered on $\boldsymbol{x}^k$, with radius $||\delta \boldsymbol{x}||_{\boldsymbol{x}=\boldsymbol{x}^k}$. The reason why we use a high dimensional ball but not the exact reachable region, i.e., $\bigcup\limits_{\boldsymbol{u}\in\boldsymbol{\mathcal{U}}} {f({\boldsymbol{x}^k},\boldsymbol{u})}$, is that $\mathcal{BA}(\boldsymbol{x}^k,||\delta \boldsymbol{x}||_{\boldsymbol{x}=\boldsymbol{x}^k})$ has a good convexity, and is computationally cheaper. Clearly, $\bigcup\limits_{\boldsymbol{u}\in\boldsymbol{\mathcal{U}}} {f({\boldsymbol{x}^k},\boldsymbol{u})}\subseteq \mathcal{BA}(\boldsymbol{x}^k,||\delta \boldsymbol{x}||_{\boldsymbol{x}=\boldsymbol{x}^k})$.

	\subsubsection{Construction of the CBF}
	In the first part, we show how to construct a candidate CBF $b(\boldsymbol{x})$ satisfying the local nonnegativity condition $\forall \boldsymbol{x}\in\boldsymbol{\mathcal{X}},0\le b(\boldsymbol{x})$ in \eqref{eq:cbf}. $b(\boldsymbol{x})$ is parameterized by:
	\begin{equation}\label{eq:paracbf}
		b(\boldsymbol{x}) = \boldsymbol{x}^\top \boldsymbol{H}_b \boldsymbol{x}+d_b,
	\end{equation}
	where $\boldsymbol{H}_b\prec 0 $, and $d_b>0$. This kind of CBF is originated from the quadratic Lyapunov function $v(\boldsymbol{x}) = -\boldsymbol{x}^\top \boldsymbol{H}_b \boldsymbol{x}$\cite{duarte2015using}, where $v(\boldsymbol{x}):\mathbb{R}^n\to\mathbb{R}$.
	Then, we have $b(\boldsymbol{x}) = d_b-v(\boldsymbol{x})$. We consider parameterizing the CBF $b(\boldsymbol{x})$ to be ellipsoidal because our application has high control frequency.
	As a result, in the state space, the Lebesgue measure of the ball $\mathcal{BA}(\boldsymbol{x}^k,||\delta \boldsymbol{x}||_{\boldsymbol{x}=\boldsymbol{x}^k})$ would be relatively small. In such a small region, the original non-linear system $\eqref{eq:kineticmodel}$ can be linearized with a slight bias.
	For a stabilizable linear system, it is reasonable to consider an ellipsoidal controlled invariant set, which is the complement set of the super-level set of an ellipsoidal Lyapunov function \cite{wang2022safe}. 
	
	With this quadratic parameterization, we can use SOS relaxation, and S-procedure \cite{derinkuyu2006s} to guarantee that $\boldsymbol{\mathcal{B}}\subseteq\boldsymbol{\mathcal{C}}$, as the controlled invariant set is inside the current reachable set:
	\begin{equation}\label{eq:sosnonnega}\tag{SOS-CBF}
		\begin{split}
			\mathrm{find}~&b(\boldsymbol{x}),\sigma_1,\nonumber\\
			\mathrm{subject~to}~&-b(\boldsymbol{x})+\sigma_1 c(\boldsymbol{x})\in\Sigma[\boldsymbol{x}],\nonumber
		\end{split}
	\end{equation}
	where $\sigma_1\in\Sigma[\boldsymbol{x}]$ is an SOS multiplier.
	We recall that the set $\boldsymbol{\mathcal{C}}$ is restricted to be a ball centered at $\boldsymbol{x}^k$, i.e. $\mathcal{BA}(x_k,||\delta \boldsymbol{x}||_{\boldsymbol{x}=\boldsymbol{x}^k})$ in the first step. Therefore, $c(\boldsymbol{x})=-||\boldsymbol{x}-\boldsymbol{x}^k||^2+||\delta \boldsymbol{x}||_{\boldsymbol{x}=\boldsymbol{x}^k}^2$. It is evident that $c(\boldsymbol{x})$ is a polynomial function, thus $\boldsymbol{\mathcal{C}}$ is a semi-algebraic set. Together with the polynomial function $b(\boldsymbol{x})$ and the SOS polynomial multiplier $\sigma_1$, the SOS constraint $\eqref{eq:sosnonnega}$ can be converted to a semi-definite constraint. We note here that the multiplier $\sigma_1$ will appear as an additional variable in the following synthesis optimization problem \eqref{eq:scensoscbf}.
	Tab.\ref{tab:set} illustrates the different sets to provide a clear understanding.
	\begin{table}[h]
		\vspace{-15pt}
		\caption{Illustration of set $\boldsymbol{\mathcal{X}}$, $\boldsymbol{\mathcal{C}}$, $\boldsymbol{\mathcal{B}}$.}
		\centering
		\begin{tabular}{lll}
			\toprule[1.5pt]
			Set & Denotation & Remark \\ \hline
			$\boldsymbol{\mathcal{X}}$ &
			safe set &
			\begin{tabular}[c]{@{}l@{}}Defined by \eqref{eq:safeset}, is not \\ explicitly applicable to the manipulator.\end{tabular} \\ \hline
			$\boldsymbol{\mathcal{C}}$ &
			reachable set &
			\begin{tabular}[c]{@{}l@{}}Defined by $\mathcal{BA}(\boldsymbol{x}^k,||\delta \boldsymbol{x}||_{\boldsymbol{x}=\boldsymbol{x}^k})$, is the \\ maximum reachable set of system \eqref{eq:kineticmodel} at $\boldsymbol{x}^k$.\end{tabular} \\ \hline
			$\boldsymbol{\mathcal{B}}$ &
			\begin{tabular}[c]{@{}l@{}}controlled\\ invariant set\end{tabular} &
			\begin{tabular}[c]{@{}l@{}}Synthesized by \eqref{eq:scensoscbf}, \\and satisfies: $\boldsymbol{\mathcal{B}}\subseteq \boldsymbol{\mathcal{C}}$, $\boldsymbol{\mathcal{B}}\subseteq \boldsymbol{\mathcal{X}}$.\end{tabular} \\ 
			\bottomrule[1.5pt]
		\end{tabular}
		\label{tab:set}
		\vspace{-10pt}
	\end{table}
	\subsection{Data-driven CBF Construction} \label{sec:datacbf}
	The following subsections show how to construct $b(\boldsymbol{x})$ for the manipulator satisfying the residual constraints in \eqref{eq:cbf} with a promising probabilistic bound. In addition to the SOS synthesis approach in the last part, we use scenario optimization to alleviate some of the constraints. The reason why SOS is not fully applicable for the remaining constraint $\forall \boldsymbol{x}\in\boldsymbol{\mathcal{C}},b(\boldsymbol{x})\le \mathrm{sd}_{\mathrm{ov}}(\boldsymbol{x})$ is that $\mathrm{sd}_{\mathrm{ov}}(\boldsymbol{x})$ is not a polynomial function in general. More precisely, there is even no explicit expression of it by hand. The constraint $\forall \boldsymbol{x}\in\partial \boldsymbol{\mathcal{C}},\exists \boldsymbol{u}\in\boldsymbol{\mathcal{U}},\frac{\partial b(\boldsymbol{x})}{\partial \boldsymbol{x}}f(\boldsymbol{x},\boldsymbol{u})\ge 0$ is also hard to convert to SOS constraints as $f(\boldsymbol{x},\boldsymbol{u})$ may not be a polynomial, and exhibits bilinearity due to the existence of control input. Although there are lifting methods \cite{anderson2015advances} and Schur relaxation methods \cite{wang2022safe} to overcome these issues, they either do not scale well with dimension or require iterative solutions. In the collision-avoidance problems for manipulators, real-time computation is rather important. This makes us turn to instead using probabilistic CBF conditions with sampled scenarios.
	
	\begin{algorithm}[t]
		\LinesNumbered
		\caption{Data-driven CBF construction}\label{al:CBF}
		\KwIn{the number of samples $\bar N$, current state $\boldsymbol{x}^k$, the maximum ball set $\mathcal{BA}(\boldsymbol{x}^k,||\delta \boldsymbol{x}||_{\boldsymbol{x}=\boldsymbol{x}^k})$.
		}
		\KwOut{CBF parameter $\boldsymbol{H}_b$ and $d_b$.
			\vspace{3pt}	
		}
		Initialize the SOS program according to \eqref{eq:scensoscbf} and $\bar N$.\\
		Randomly generate $\bar N$ samples according to $\pi(\boldsymbol{x})$\\
		\For{$i\leq\bar{N}$}{
			Compute $\mathrm{sd}_{\mathrm{ov}}(\boldsymbol{x}^{(i)})$ for all samples in $\bar {X}$.\\
		}
		Solve the optimal problem \eqref{eq:scensoscbf}.\\
		Return $\boldsymbol{H}_b$ and $d_b$.\\
		
	\end{algorithm}
	
	The scenario optimization relies on sampled scenarios to relax the original problem. 
	Sampling all SDF with $\boldsymbol{x}\in\boldsymbol{\mathcal{C}}$ and solving \eqref{eq:sosnonnega} is impractical since the number of samplings is infinite.
	Instead, we sample finite $\bar N$ realizations of $\boldsymbol{x}^{(r)}$ around $\boldsymbol{x}^k$.
	The samples are with a probability measure $\pi$, which satisfies:
	\begin{equation}\label{eq:probability}
		\int_{\boldsymbol{\mathcal{C}}} {\pi(\boldsymbol{x}) d\boldsymbol{x}}=1.
	\end{equation}

	Let $\bar{ \boldsymbol{X}}=\{\boldsymbol{x}^{(1)},\boldsymbol{x}^{(2)},\ldots,\boldsymbol{x}^{(\bar N)}\}$ be the set of SDF sampled scenarios. These scenarios are independently and identically sampled according to $\pi$. We could construct the following scenario program:
	\begin{subequations}\label{eq:scencbf}
		\begin{align}
			\mathrm{find}_{\boldsymbol{u}\in\boldsymbol{\mathcal{U}}}~&b(\boldsymbol{x}),\boldsymbol{u} \label{eq:scencbf-1}\\
			\mathrm{subject~to}~&b(\boldsymbol{x}^{(i)})\le \mathrm{sd}_{\mathrm{ov}}(\boldsymbol{x}^{(i)}),\label{eq:scencbf-2}\\
			&\left.\frac{\partial b(\boldsymbol{x})}{\partial \boldsymbol{x}}f(\boldsymbol{x},\boldsymbol{u})+\alpha b(\boldsymbol{x})\right|_{\boldsymbol{x}=\boldsymbol{x}^{(i)}}\ge 0,\label{eq:scencbf-3}\\
			&\forall~\boldsymbol{x}^{(i)}\in\bar{ \boldsymbol{X}}.\label{eq:scencbf-4}
		\end{align}
	\end{subequations}
	The lower-envelope condition $\forall \boldsymbol{x}\in\boldsymbol{\mathcal{C}},b(\boldsymbol{x})\le \mathrm{sd}_{\mathrm{ov}}(\boldsymbol{x})$ is enforced only on the finite set of scenarios $\bar{ \boldsymbol{X}}$. The controlled invariance condition \eqref{eq:boundary} is substituted by a relaxed formulation $\forall \boldsymbol{x}\in \boldsymbol{\mathcal{B}},\exists \boldsymbol{u}\in\boldsymbol{\mathcal{U}},\frac{\partial b(\boldsymbol{x})}{\partial \boldsymbol{x}}f(\boldsymbol{x},\boldsymbol{u})+\alpha b(\boldsymbol{x})\ge 0$, where $\alpha\in\mathbb{R}_+$. This relaxed formulation leads to a convex problem, which helps with finding a numerical solution.
	The additional term $\alpha b(\boldsymbol{x})$ is motivated by the zero-CBF approach \cite{ames2016control}. 
	The final synthesis program for a data-driven CBF with SDF samples for manipulators is given as follows:
	\begin{theorem}
		For dynamical system \eqref{eq:kineticmodel}, sampling a set of scenarios $\bar{ \boldsymbol{X}}=\{\boldsymbol{x}^{(1)},\boldsymbol{x}^{(2)},\ldots,\boldsymbol{x}^{(\bar N)}\}\in\boldsymbol{\mathcal{C}}$, the CBF $b(\boldsymbol{x})$ at the current state $\boldsymbol{x}^{k}$ can be constructed by solving the following program:
		\begin{equation}\tag{SCSOS-CBF}\label{eq:scensoscbf}
			\begin{split}
				\mathop{\max}\limits_{\boldsymbol{u}\in\boldsymbol{\mathcal{U}},\sigma_1\in\Sigma[\boldsymbol{x}],\boldsymbol{H}_b\prec 0,d_b>0}~&d_b\\
				\mathrm{subject~to}~&\eqref{eq:scencbf-2},\eqref{eq:scencbf-3},\eqref{eq:scencbf-4},\\
				&-b(\boldsymbol{x})+\sigma_1 c(\boldsymbol{x})\in\Sigma[\boldsymbol{x}],\\
				&||\boldsymbol{H}_b||_2=1.
			\end{split}
		\end{equation}
	\end{theorem}
	\begin{proof}
		Condition $\mathrm{sd}_{\mathrm{ov}}(\boldsymbol{x}^{(i)})\le b(\boldsymbol{x}^{(i)})$ indicates that the control invariant set is a subset of the safe set in the sense of scenarios $\boldsymbol{x}^{(i)}\in\bar{ \boldsymbol{X}}$. The condition $\frac{\partial b(\boldsymbol{x})}{\partial \boldsymbol{x}}f(\boldsymbol{x},\boldsymbol{u})+\alpha b(\boldsymbol{x})\ge 0$ leads to a convex problem when seeking barrier functions with numerical means. Condition $\forall~\boldsymbol{x}^{(i)}\in\bar{ \boldsymbol{X}}$ restricts the sampling points in the reachable set $\boldsymbol{\mathcal{C}}$.
		Condition $-b(\boldsymbol{x})+\sigma_1 c(\boldsymbol{x})\in\Sigma[\boldsymbol{x}]$ indicates that for any $\boldsymbol{x}$, $-b(\boldsymbol{x})+\sigma_1 c(\boldsymbol{x}) \geq 0$ and further $\forall~\boldsymbol{x}\in\boldsymbol{\mathcal{B}}$, $c(\boldsymbol{x})\geq 0$.
		Regularization of $\boldsymbol{H}_b$ enables the volume reservation of the set $\boldsymbol{\mathcal{B}}$ by maximizing $d_b$.
	\end{proof}
	
	Here we note that our method does not require parameterizing the controller, unlike the results in literature \cite{wang2022safe}. The objective is to maximize the value of $d_b$ to reserve the volume of the control invariant set $\boldsymbol{\mathcal{B}}$, under the regularization of $\boldsymbol{H}_b$. 
	Thus, manipulators can be controlled in a cluttered environment efficiently.
	The decision variables $\boldsymbol{u},\sigma_1,\boldsymbol{H}_b,d_b$ are stacked by $z\in\mathbb{R}^{e}$ for the ease of following theoretic analysis, where $e=m+1+n^2+1$. The set of active constraints \emph{about scenarios} is defined by $\mathcal{S}_{\boldsymbol{x}}$. 

	Algorithm \ref{al:CBF} is a data-driven methodology for CBF synthesis considering the manipulator dynamics.
	Once the SOS program is built, we can solve \eqref{eq:scensoscbf} with a semi-definite programming solver.
	When the next state $\boldsymbol{x}^{k+1}$ comes in, we only go through lines 2-7 in Algorithm \ref{al:CBF}.
	
	\subsection{Probabilistic Guarantee for Safety}
	Since the number of sampling is finite, the data-driven method cannot guarantee the equivalence between $b(\boldsymbol{x}^{(i)})\le \mathrm{sd}_{\mathrm{ov}}(\boldsymbol{x}^{(i)})$ and $b(\boldsymbol{x})\le \mathrm{sd}_{\mathrm{ov}}(\boldsymbol{x})$.
	Assume that $z$ is the solution of \eqref{eq:scensoscbf}.
	It would be possible that $z$ is not in the safe solution set $\mathcal{S}_x$ and the manipulator is controlled to collision, due to the uncertainty caused by sampling.
	The \emph{violation probability} is given as:
	\begin{definition}[violation probability \cite{campi2008exact}]\label{def:violation}
		The \emph{violation probability} of a given solution $z$ is defined as $V(z)=\mathbb{P}\{\boldsymbol{x} \in \boldsymbol{\mathcal{C}}: z \notin \mathcal{S}_x \}$.
	\end{definition}
	Recent results point out that the violation probability $V(z)$ is closely related to both the number of scenarios and the \emph{complexity}, i.e. number of \emph{support constraints} \cite{garatti2019risk}.
	\begin{algorithm}[htb]
		\LinesNumbered
		\caption{Determine the number of samples}\label{al:probability}
		\KwIn{the dimension of the decision variables $e$, confidence parameter $\beta$, risk level $\epsilon$, the predefined maximum number $\bar{N}_{\max}$, probability threshold $\epsilon_p$.
		}
		\KwOut{the minimum number of the sample $\bar{N}$.
			\vspace{3pt}	
		}
		$\bar{N}_{\min} = e$.\\
		\While{$\bar{N}_{\min}+1 \le \bar{N}_{\max}$}
		{
			compute the inverse incomplete beta function $\to \gamma_L$ and $\to \gamma_U$, based on $(\beta, e, \bar{N}_{\min}-e+1)$ and $(\beta, e, \bar{N}_{\max}-e+1)$, respectively.\\
			$t_{L1} = 1-\gamma_L$,
			$t_{L2} = 1$,
			$t_{U1} = 1-\gamma_U$,
			$t_{U2} = 1$.\\
			obtain $P_{L1}$ $P_{L2}$ $P_{U1}$ and $P_{U2}$ according to \eqref{eq:scenequation} based on $t_{L1}$, $t_{L2}$, $t_{U1}$ and $t_{U2}$.\\
			\eIf{$P_{L1}\times P_{L2} \ge 0$}
			{$\epsilon_1 = 0$.}
			{
				\While{$t_{L2}-t_{L1} > 0$}{
					$t_L = \lceil(t_{L2}+t_{L1})/2\rceil$.\\
					obtain $P_{tL}$ according to \eqref{eq:scenequation} based on $t_L$.\\
					\eIf{$P_{tL}>0$}{$t_{L1} = t_L$.}{$t_{L2} = t_L$.}
					$\epsilon_1 = 1-t_{L2}$.
				}
			}
			compute $\epsilon_2$ based on $t_{U1}$, $t_{U2}$, $P_{U1}$ and $P_{U2}$.\\
			\eIf{$|\epsilon_1-\epsilon| > \epsilon_p$ or $|\epsilon_1-\epsilon_2| > \epsilon_p$}{$\bar{N}_{\min} = \bar{N}_{\min} + \lceil (\bar{N}_{\min} - \bar{N}_{\max})/2 \rceil$}{$\bar{N} = \bar{N}_{\min}$.\\
				return $\bar{N}$}
			
		}
	\end{algorithm}
	\begin{definition}\label{def:support}
		A constraint in $\mathcal{S}_x$ of the synthesis program \eqref{eq:scensoscbf} is called a \emph{support constraint} if its removal (while all the other constraints are maintained) changes the optimal solution. The \emph{complexity} $c_{\bar N}^*$ of the synthesis scenario program \eqref{eq:scensoscbf} is the number of the support constraints. 
	\end{definition}
	Then, the probabilistic result of violation probability based on the $\bar{N}$ scenarios is given in the following results.
	
	\begin{theorem}[adopted from \cite{garatti2019risk}]\label{th:probability}
		Consider the CBF synthesis program \eqref{eq:scensoscbf} with $\bar N>e=m+n^2+2$. Given confidence parameter $\beta\in(0,1)$, for any $k=0,1,\ldots,e$, consider the polynomial equation in the $\xi$ variable
		\begin{equation}\label{eq:scenequation}
			\left(
			\begin{aligned}
				{\bar N}\\
				k
			\end{aligned}
			\right){\xi^{\bar N - k}} - \frac{\beta }{{2\bar N}}\sum\limits_{i = k}^{\bar N - 1} {\left( \begin{aligned}
					i\\
					k
				\end{aligned}\right){\xi^{i - k}}\\
				= \frac{\beta }{{6\bar N}}\sum\limits_{i = \bar N + 1}^{4\bar N} {\left( \begin{aligned}
						i\\
						k
					\end{aligned} \right){\xi^{i - k}} } }. 
		\end{equation}
		
		This equation has two solutions in $[0,+\infty)$ which are denoted by $\underline \xi(k)$ and $\bar \xi(k)$, where $\underline \xi(k)\le \bar \xi(k)$. Let $\underline\epsilon(k):=\max\{0,1-\bar \xi(k)\}$ and $\bar \epsilon:=1-\underline \xi(k)$. Then, for $\boldsymbol{\mathcal{C}}$ and any $\pi$, it holds that
		\begin{equation}\label{eq:probability1}
			\mathbb{P}^{\bar N}\{\underline\epsilon(c_{\bar N}^*)\le V(z^*)\le \bar\epsilon(c_{\bar N}^*)\}\ge 1-\beta.
		\end{equation}
	\end{theorem}
	This result shows the relationship between the number of support constraints $c_{\bar N}^*$, the violation probability on the optimal solution $V(z^*)$ and parameter $\beta$. The scenario constraints are more prone to be violated if the complexity is high. One intuitive interpretation of this result is that the higher complexity is, the more boundary of constraints the sorted solution stands on. Then, the uncertain constraints have a higher risk. The following corollary is a direct result from Theorem \ref{th:probability}.
	\begin{corollary}
		Given $\beta$, it always holds that $\underline\epsilon(c_{\bar N}^*)\le \underline\epsilon(e)\le \bar \epsilon(c_{\bar N}^*)\le \bar \epsilon(e)$. Besides, we certainly have $\mathbb{P}^{\bar N}\{V(z^*)\le \bar \epsilon(c_{\bar N}^*)\}\ge 1-\beta$.
	\end{corollary}
	One direct application of this result is that we can measure at most how many samples are required for a given confidence parameter $\beta$ and risk level $\epsilon$ for manipulators. The following lemma concludes the amount of data.
	\begin{lemma}\label{lem:amount}
		Given risk level $\epsilon\in[0,1)$, confidence parameter $\beta\in[0,1)$, then the amount $\bar N(\epsilon,\beta)$ of samples required to render $\mathbb{P}^{\bar N}\{V(z^*)\le\bar \epsilon\}\ge 1-\beta$ fulfills:
		\begin{equation}\label{eq:amouont}
			\bar N(\epsilon,\beta) \ge \left\{ \begin{array}{l}
				\mathop {\arg \min }\limits_{\bar N \in \mathbb{N}}~\bar N\\
				\mathrm{s.t. }\eqref{eq:scenequation}
			\end{array} \right.
		\end{equation}
	\end{lemma}
	Although Lemma \ref{lem:amount} gives guidance on how many sampled data are necessary for the acceptable risk level and confidence, the result is hard to obtain since the optimization problem in \eqref{eq:amouont} is a non-convex mixed integer program. We provide a heuristic algorithm which can compute $\bar{N}$ given $e$, $\epsilon$ and $\beta$. There are two levels of the dichotomic search program in 
	Algorithm \ref{al:probability}. The first one computes the minimum number of samples. The second one computes the risks. Line 21 gives the terminal conditions: 1) the risk based on $\bar{N}$ samples is near the risk goal, 2) the risk cannot decrease too much with $\bar{N}$ increasing.

	\section{Simulation and Experimental Implementation}\label{sec:sim}
	This section gives the control synthesis with the proposed CBF construction to formulate a safety-critical control for the robotic manipulator.
	\subsection{Control Synthesis}
	We pre-construct a trajectory as a series of way-points by planning methods for the manipulators, but it is not necessarily safe. $\boldsymbol{u}_{des}$ in \eqref{eq:SCB-QP} is computed with a P controller to the next way-point.
	Similar to \cite{singletary2022safety}, we also set a way-point switch mechanism to avoid the manipulator getting stuck.
	As pointed out in \cite[Proposition 1]{singletary2022safety}, \eqref{eq:SCB-QP} can guarantee safety with the kinematic model of robotic manipulators when using an exponential stable low-level velocity tracking controller.
	Specifically, the states $\boldsymbol{x}$ are the configuration of each joint, and the inputs $u_k$ are the velocity of each joint.
	
	\begin{definition}[Low-level velocity tracking controller]
		For a velocity command $v_c(\boldsymbol{x}; t)$, consider the corresponding tracking error as $	\dot{\boldsymbol{\varepsilon}} = \dot{\boldsymbol{x}}-v_c(\boldsymbol{x}; t)$.
		The low-level velocity tracking controller $\boldsymbol{u} = k(\boldsymbol{x}; t)$ can exponentially stabilize the tracking as $	\|\dot{\boldsymbol{\varepsilon}} (t)\|_2\leq l {\boldsymbol{\varepsilon}} ^{-\lambda t}\|\dot{\boldsymbol{\varepsilon}} (t_0)\|$,
		where $l,\lambda>0$.
	\end{definition}

	The collision-free behavior is enforced for the kinematic model of the manipulator by constructing the CBF $b(\boldsymbol{x})$ online through Algorithm \ref{al:CBF}, and solving \eqref{eq:SCB-QP}.
	\begin{figure}[t]
		\centering
		\includegraphics[width=1\linewidth]{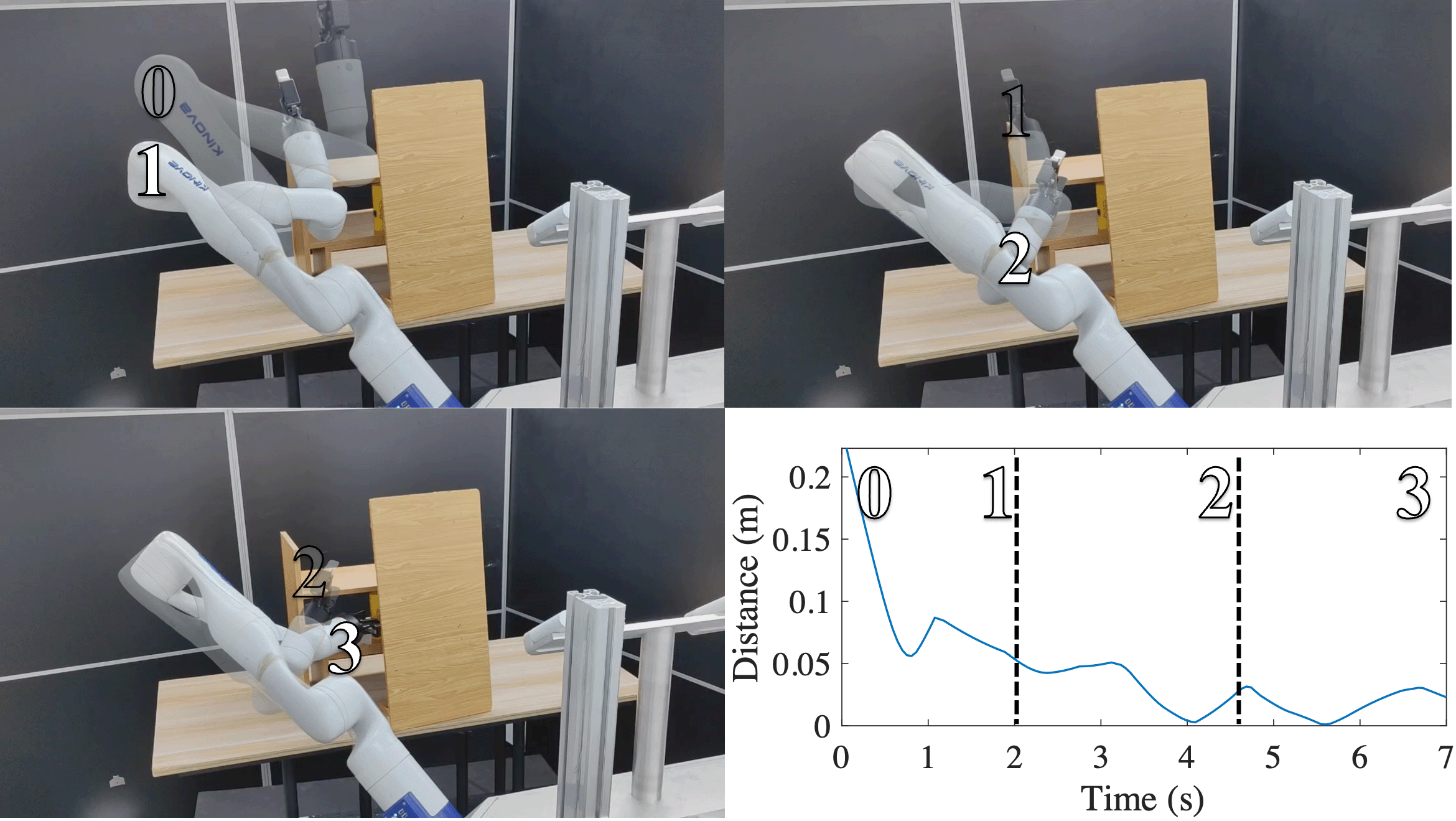}
		\caption{Experiment implemented on the Kinova Gen3.
			The manipulator reaches an object on the left side.
			The bold Arabic numerals stand for different phases of the obstacle avoidance task.
			0: the initial pose; 1: move towards the object; 2: avoid the board; 3: reach the object.
			The smallest distance between the manipulator and the obstacle is about one millimeter.
		}
		\label{fig:scene1}
	\end{figure}
	
	\begin{table}[b]
		\vspace{-5pt}
		\caption{Theoretical risk and computation times of the data-driven CBF-construction}
		\centering
		\begin{tabular}{ccc}
			\toprule[1.5pt]
			Theoretical risk & Number of   samplings & Computation time (ms) \\
			\midrule[0.75pt]
			0.5              & 108                   & 27.41                 \\
			0.4              & 138                   & 27.98                 \\
			0.3              & 188                   & 25.25                 \\
			0.2              & 288                   & 24.49                 \\
			0.1              & 588                   & 27.55                 \\
			0.05             & 1188                  & 28.28                \\
			\bottomrule[1.5pt]
		\end{tabular}
		\label{tb:time1}
	\end{table}
	\subsection{Implementation}
\begin{figure}[t]
		\centering
		\includegraphics[width=1\linewidth]{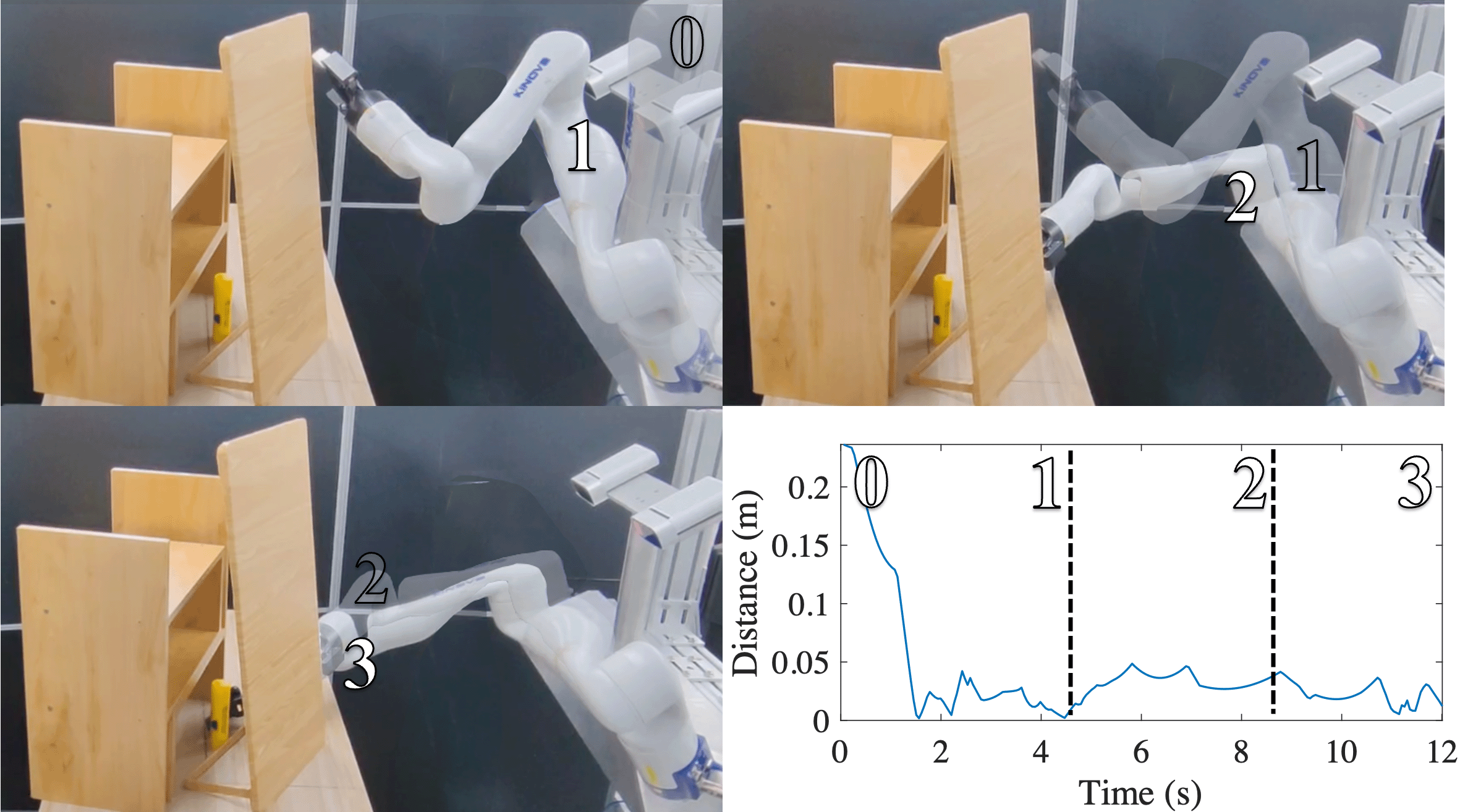}
		\caption{Experiment implemented on the Kinova Gen3.
			The manipulator reached an object on the right side.
			0: the initial pose; 1: move towards the board; 2: move across the board; 3: reach the object.
			The smallest distance between the manipulator and the obstacle is about two millimeters.
		}
		\label{fig:scene2}
	\end{figure}

	We implement our methodology for manipulator motion planning in obstacle-cluttered environments to validate the efficacy.
	The manipulator, obstacles, and objects are a series of fine-shaped meshes (0.02 mm tolerance).
	Furthermore, the manipulator is described in a Unified Robot Description Format (URDF) in the simulation environment with \textit{Robotics Toolbox} in MATLAB. 
	Two real experimental scenarios (see Figs.\ref{fig:scene1} and \ref{fig:scene2}).
	We aim to use a Kinova Gen 3 robotic manipulator to grasp an object behind a board on a shelf.
	Our method first requires determining the number of samples needed to construct \eqref{eq:scensoscbf} by Algorithm 2.
	Then, \eqref{eq:scensoscbf} is solved with \textit{Sedumi} and \textit{SOSTOOL}\cite{papachristodoulou2013sostools} based on Algorithm 1.
	Finally, \eqref{eq:SCB-QP} is solved, and the command is sent to the manipulator.
	All these procedures are carried out on a computer with an Intel(R) Core(TM) i9-9980XE CPU, 3.00GHz processor, and 64GB RAM.

	In the real scene, the primary obstacle of concern is the shelf made of eight boards and one block on the desk.
	The object is a solid glue behind the block on the shelf.
	This scene stands for a typical application where the manipulator tends to grasp something in a complicated and cluttered indoor environment.
	For example, use a manipulator to grab and unplug a charger behind an LCD monitor. 
	When performing such tasks, the clearance between the manipulator and the obstacles is less than a few millimeters.
	Our methods can achieve such tasks since the SDF is not linearized, thus enlarging the volume of the feasible space.
	The data-driven CBF further guarantees safety for model \eqref{eq:kineticmodel}.
	Figures \ref{fig:scene1} and \ref{fig:scene2} show the motions and distances throughout the experiment.
	\begin{figure}[t]
		\centering
		\includegraphics[width=1\linewidth]{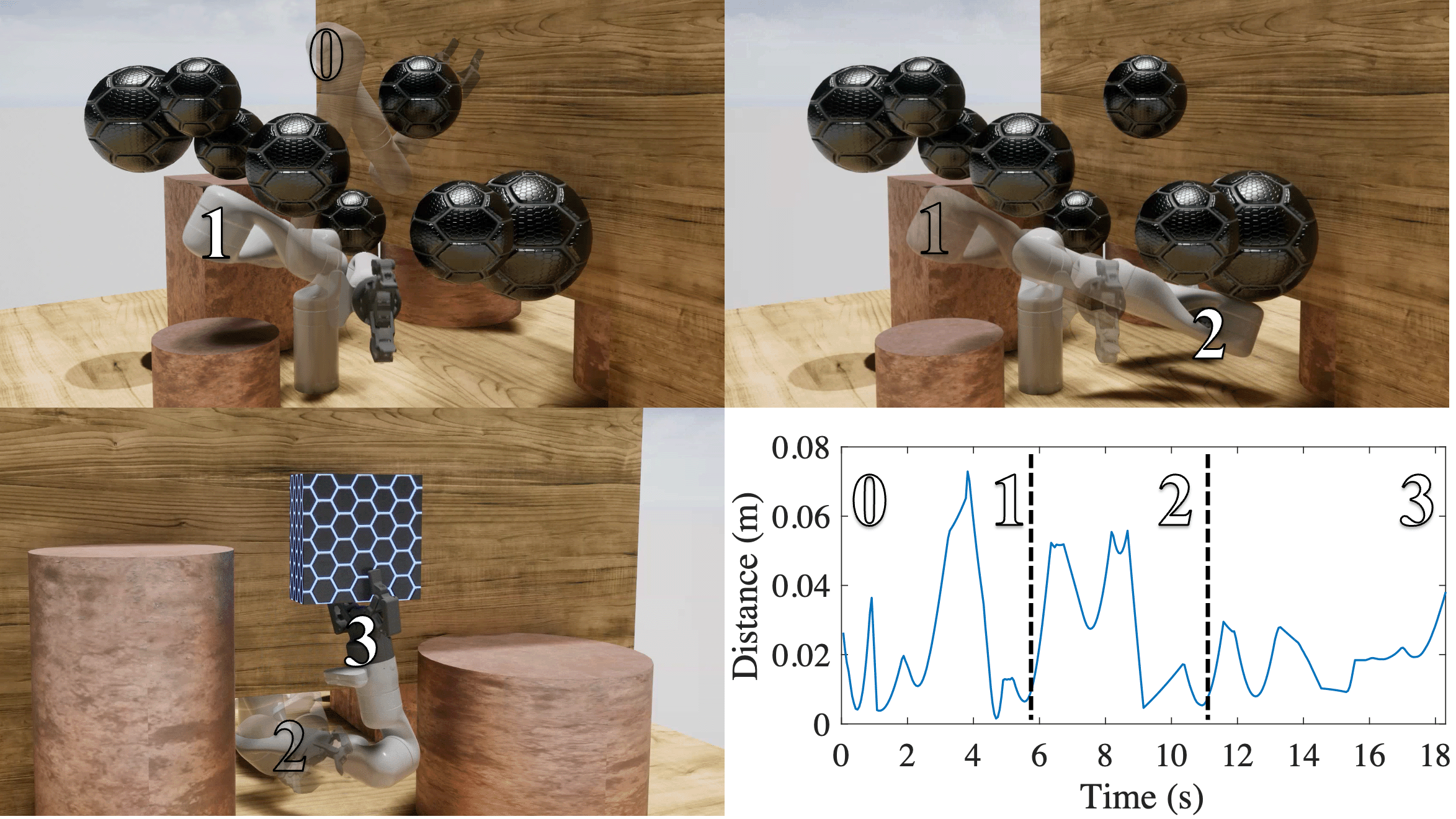}
		\caption{Experiment implemented on the Kinova Gen3 in a virtual environment with static obstacles.
			0: the initial pose; 1: move downwards; 2: move through the window; 3: reach the object.
			The smallest distance between the manipulator and the obstacle is about four millimeters.
		}
		\label{fig:vscene1}
	\end{figure}
	
\begin{figure}[t]
		\centering
		\includegraphics[width=1\linewidth]{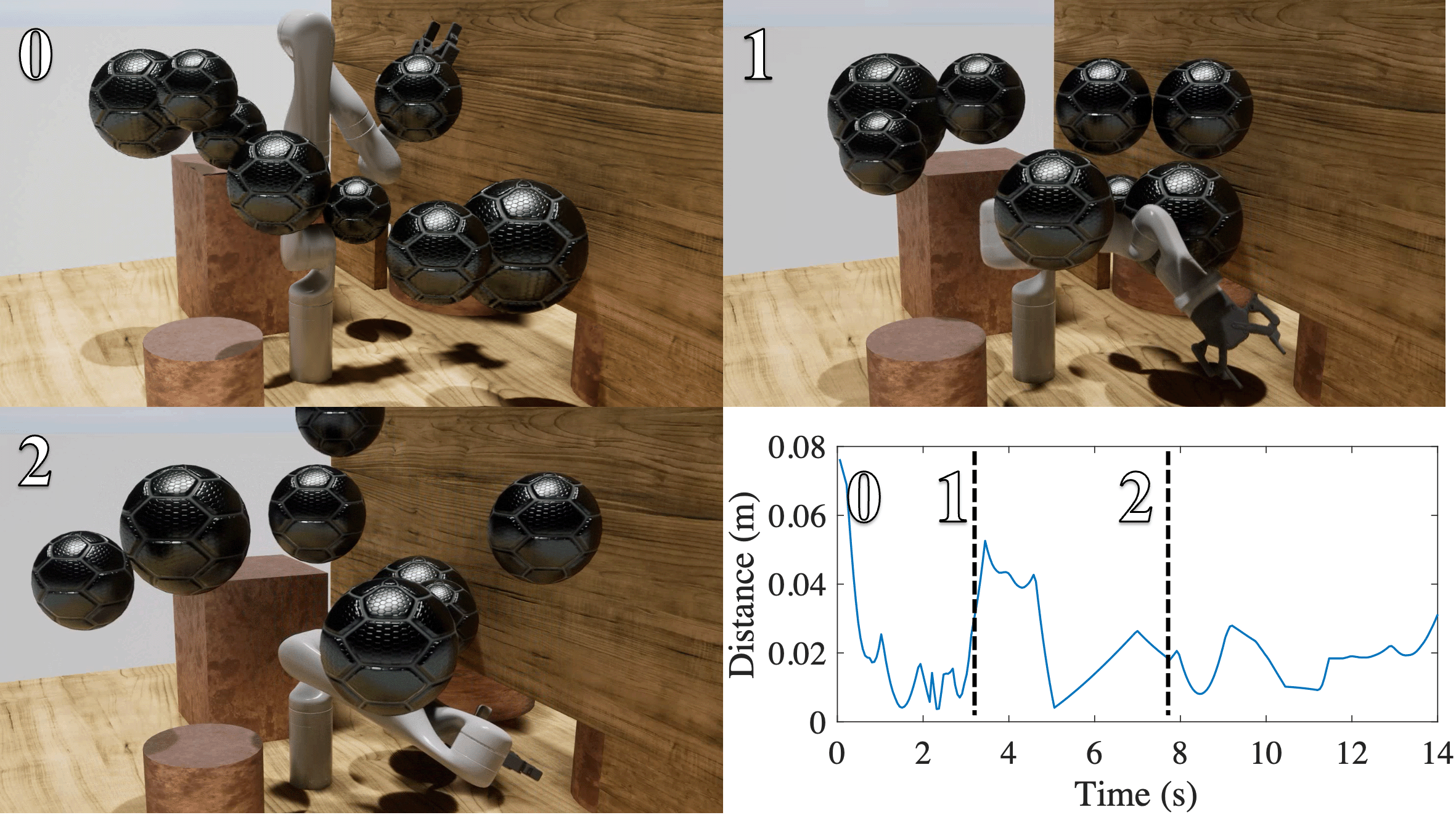}
		\caption{Experiment implemented on the Kinova Gen3 in a virtual environment with moving obstacles. The SDF is sampled online.}
		\label{fig:vscene2}
	\end{figure}

	We then design a much more complex environment in the virtual scene to test the method.
	The obstacles of concern are eight balls, four pillars, and one giant board.
	The object is on the other side of the board.
	The manipulator must pass through the window at the bottom of the board while avoiding the collision.
	Our method can achieve this task with a clearance of about four millimeters.
	Figure \ref{fig:vscene1} shows the virtual motion and the distance throughout the trajectory.
	We continued to simulate the virtual scene where the eight balls were moving.
	Figure \ref{fig:vscene2} shows the virtual test with moving obstacles and the distance throughout the motion.
	The video can be found at \url{https://www.youtube.com/watch?v=io0sXuJfvAE}.
	
	Each scene is tested ten times with different numbers of samples.
	Consequently, the theoretical safety probability is quite different, but
	\textit{all the tests are successfully operated with safety in practice.}
	In algorithm 2, we set the $\beta = 0.05$, $\epsilon$ from $0.5$ to $0.05$.
	The computation time of our data-driven CBF-construction method is shown in tab.\ref{tb:time1}.
	The theoretical risk decreases as the sampling number grows.
	Using 1188 samples, it has a 95\% probability that the task is safe.
	The computation time does not increase with the number of sampling, which shows that our method is computationally efficient.
	The relaxation in \eqref{eq:scensoscbf} guarantees that our method has real-time ability.
	As for the average QP computation time, it is 12.41 ms.
	For comparison purposes, we also conducted the test based on CBF \cite{singletary2022safety} and TrajOpt.
	The average computation time for QP in CBF is 12.34 ms, which is similar to our result.
	The average computation time for TrajOpt is 270 ms.
	The CBF and TrajOpt cannot be successful in all tests, especially when the clearance is within millimeters.
	The average computation time for our method is 39.31 ms.
	
	\section{Conclusion}\label{sec:con}
	In this paper, a data-driven CBF construction method is proposed. The method tails the characteristic of the system model to ensure safety under the merit of control invariance. By sampling data in real-time and incorporating the samplings as scenarios in the optimization problem, our method provides less conservative results with a probability guarantee. We also validate the proposed algorithm on industrial robotic manipulators to perform several safety-critical control tasks. In the future, we will explore cooperative assembly tasks by a multi-manipulator system.

	\bibliographystyle{ieeetr}
	\bibliography{ref.bib}
	
\end{document}